\documentclass[a4paper]{article}

\usepackage{INTERSPEECH2019}
\usepackage{cite}
\usepackage{tikz}
\usepackage{url}
\usepackage{multirow}
\usepackage{graphicx}
\usepackage[prependcaption,textsize=scriptsize]{todonotes}
\usepackage{xcolor}

\title{Improved low-resource Somali speech recognition by semi-supervised acoustic and language model training} 
\name{Astik Biswas, Raghav Menon, Ewald van der Westhuizen, Thomas Niesler}
\address{Department of Electrical and Electronic Engineering, Stellenbosch University, South Africa}
\email{abiswas@sun.ac.za, rmenon@sun.ac.za, ewaldvdw@sun.ac.za, trn@sun.ac.za}

\begin{document}

\maketitle
\begin{abstract}

We present improvements in automatic speech recognition (ASR) for Somali, a currently extremely under-resourced language.
This forms part of a continuing  United Nations (UN) effort to employ ASR-based keyword spotting systems to support humanitarian relief programmes in rural Africa.
Using just 1.57 hours of annotated speech data as a seed corpus, we increase the pool of training data by applying semi-supervised training to 17.55 hours of untranscribed speech.
We make use of factorised time-delay neural networks (TDNN-F) for acoustic modelling, since these have recently been shown to be effective in resource-scarce situations.
Three semi-supervised training passes were performed, where the decoded output from each pass was used for acoustic model training in the subsequent pass.
The automatic transcriptions from the best performing pass were used for language model augmentation.
To ensure the quality of automatic transcriptions, decoder confidence is used as a threshold.
The acoustic and language models obtained from the semi-supervised approach show significant improvement in terms of WER and perplexity compared to the baseline.
Incorporating the automatically generated transcriptions yields a 6.55\% improvement in language model perplexity.
The use of 17.55 hour of Somali acoustic data in semi-supervised training shows an improvement of 7.74\% relative over the baseline.

\end{abstract}
\noindent\textbf{Index Terms}: speech recognition, Somali, semi-supervised, TDNN-F, under-resourced language


\section{Introduction}
In countries with a well established internet infrastructure, social media has become an accepted platform for sharing opinions and concerns~\cite{Vosoughi_ICDMW15, Wegrzyn_CASoN11, Burnap15}. 
Surveys conducted by the United Nations (UN) in places lacking sufficient internet infrastructure indicate that this function is fulfilled by radio phone-in shows~\cite{unpulse1, unpulse2, unpulse3}.
Therefore, to support its humanitarian relief efforts in rural and under-developed parts of Uganda, the UN has piloted radio browsing systems in three of the country's languages ~\cite{Menon2017, Saeb2017}.
 
The success in Uganda served as a motivator for the development of a corresponding system for Somalia, an African country where the UN is also currently engaged.
However, the Somali language is extremely under-resourced and the difficulty of compiling speech data resulted in an available training set of just 1.57 hours~\cite{menon2018automatic}.
By leveraging available resources from better-resourced but unrelated languages~\cite{biswas2018IS, ghoshal2013multilingual, Saeb2017}, a system using a hybrid neural network acoustic model was able to achieve a word error rate (WER) of 53.75\% \cite{menon2018automatic}.
While the additional language resources were beneficial to the acoustic model, we found that care had to be taken when deciding on the composition of the multilingual training set.
To our knowledge, only one other study on Somali automatic speech recognition (ASR) has so far been described in the literature~\cite{Nimaan2006}.

Somali is an Afroasiatic language.
It is the official language of Somalia and widely used its neighbouring countries.\footnote{\url{https://www.alsintl.com/resources/languages/Somali/}}
There are an estimated 7 million native Somali speakers in Somalia and 10 to 16 million worldwide.
Somali is written using the Latin alphabet, and has a phoneme inventory of 23 consonants and five vowels.
Somali is an agglutinative language that is characterised by a high number of morphemes per word.

The multilingual Somali acoustic model described in \cite{menon2018automatic} uses a hybrid neural network that contains several million parameters.
This complexity increases the computational demand of the decoding process.
Since the computational resources available in the target setting are very limited, it is important to reduce the required computation, for example by using an acoustic model with fewer parameters.
The recently-introduced factorised time-delay neural networks (TDNN-F)~\cite{povey2018} utilise half the number of parameters than the hybrid networks with comparable performance, in particular in a low-resource setting.
This motivated us to consider this neural network architecture for acoustic modelling using our extremely small Somali training corpus.

In the following, we present the results of our most recent efforts to improve our Somali ASR system.
We make use of TDNN-F acoustic models and experiment with the incorporation of additional but unannotated Somali speech data by semi-supervised training, an approach which has been applied successfully in some other low-resource settings~\cite{yilmaz2018semi,nallasamy2012semi,Saeb2017,thomas2013deep}.  


\vspace{-6pt}
\section{Radio browsing system}
\label{rbs}

\begin{figure}[ht]
  \centering
  \includegraphics[width=60mm]{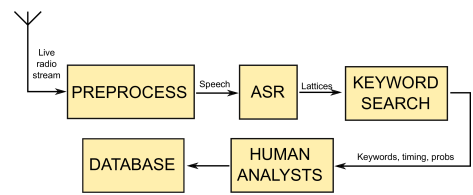}
  \caption{\it The radio browsing system~\cite{Saeb2017}.}
  \label{fig:radio_browsing_system}
  \vspace{-14pt}
\end{figure}

Figure~\ref{fig:radio_browsing_system}~\cite{Saeb2017} shows the components of the radio browsing system.
The preprocessed audio stream is passed to the ASR system which generates lattices which are subsequently searched for predefined keywords. 
Human analysts further process the data which aid in humanitarian decision making and situational awareness. 
This system is currently successfully deployed by the UN in Uganda.\footnote{Examples of system output are available at~\mbox{\url{http://radio.unglobalpulse.net}}
}

\vspace{-6pt}

\section{Acoustic and text data}

\subsection{Manually transcribed acoustic data}
\label{sec:transcribedacousticdata}
The Somali acoustic training and test data used in our experiments is described in Table~\ref{tab:acousticdatasomali}.
This small dataset of speech captured from broadcast Somali radio phone-in programmes, contains only 1.57 hours of transcribed speech that is available for training and 10 minutes for testing.

 
\begin{table}[!ht]
\vspace{-8pt}
\small
\centering
\caption{Somali transcribed speech data. Duration is indicated in hours (h) and minutes (m).}
\vspace{-6pt}
\renewcommand*{\arraystretch}{0.8}
\label{tab:acousticdatasomali}
\begin{tabular*}{0.47\textwidth}{@{\extracolsep{\fill}}lrr}
\toprule
\textbf{Speech dataset} & \textbf{Utterances} & \textbf{Duration} \\
\midrule
Training set    &  1.3k & 1.57h \\
Test set        &    95 & 10m   \\ \midrule
Total           &  1.4k & 1.74h \\  \bottomrule
\end{tabular*}%
\vspace{-8pt}
\end{table}

Besides the Somali data, the larger datasets used to train the Ugandan systems were available for acoustic modelling, as shown in Table~\ref{tab:acousticdataother}.
Luganda and Acholi are indigenous Ugandan languages, while Ugandan English is highly accented.
In addition to the Ugandan data, a 20-hour dataset of South African English Broadcast News was available.
While the Ugandan data was drawn exclusively from radio phone-in talk shows, the South African data was compiled from national radio news bulletins and consists of a mix of prepared and spontaneous speech~\cite{Kamper2015}.
The total transcribed multilingual speech data available for acoustic model training (ManT), comprising Somali and these additional languages, adds up to 46.37 hours. 

\begin{table}[!ht]
\small
\centering
\renewcommand*{\arraystretch}{0.8}
\caption{Transcribed speech data from other languages.}
\vspace{-8pt}
\label{tab:acousticdataother}
\begin{tabular*}{0.47\textwidth}{@{\extracolsep{\fill}}lrr}
\toprule
\textbf{Speech dataset} & \textbf{Utterances} & \textbf{Duration} \\
\midrule
Luganda              &  8.8k & 9.6h \\
Acholi                &  4.9k & 9.2h \\
Ugandan English       &  4.4k     & 6.0h     \\
South African English &   10.5k    & 20.0h     \\
\midrule
Total                 & 28.6k & 44.8h \\
\bottomrule
\end{tabular*}%
\vspace{-8pt}
\end{table}


\subsection{Untranscribed acoustic data}
\label{sec:untranscribedacousticdata}
Approximately 17.55 hours of untranscribed Somali speech data, also collected from phone-in programmes, was available for experimentation.
No information regarding speaker identity or other characteristics of the speech was available.
As a simple and na\"ive first approach, the speech files were simply divided into fixed-length segments before being passed to the recogniser for transcription and semi-supervised retraining.
In future, the effect of more sophisticated segmentation, based for example on diarisation, will be investigated.
The automatically transcribed data, or specific subset thereof, that is output by the transcriptions systems will be referred to as `AutoT.'

\subsection{Text data}
Table~\ref{tab:textdata} summarises Somali text corpora that were available for language modelling.
Besides the 15k words in the transcriptions of the Somali speech corpus training set (Table~\ref{tab:acousticdatasomali}), a number of additional resources were available.
Approximately 2M words were harvested from publicly-accessible news websites.
A further 1.6M words were gathered from public Facebook posts and 3.5M words from selected Facebook comments. 
Comments were replies to the posts and were generally less well edited.
While the Somali news text and Facebook posts were carefully manually cleaned and filtered, the Facebook comments consist of raw, unfiltered text~\cite{menon2018automatic}.
Furthermore, two corpora taken from the Leipzig Corpora Collection (LCC)~\cite{lcc2012building} were included in our language model (LM) data collection.
Finally, it has been shown by some researchers that text generated artificially using a long short-term memory (LSTM) neural network can lower language model perplexity when incorporated as additional training data~\cite{yilmaz2018semi}.
Hence we trained an LSTM network on the Somali acoustic training transcriptions and generated an 11M word corpus of artificial text.

\begin{table}[h]
\footnotesize
\centering
\renewcommand*{\arraystretch}{0.8}
\caption{Somali text resources used for language modelling.}
\vspace{-8pt}
\label{tab:textdata}
\begin{tabular*}{0.47\textwidth}{@{\extracolsep{\fill}}l@{\hspace{3pt}}l@{\hspace{3pt}}r@{\hspace{6pt}}r@{\hspace{6pt}}r}
\toprule
 & \textbf{Corpus} & \textbf{Word tokens} & \textbf{Word types} & \textbf{Sentences} \\
\midrule
T1 & Somali transcriptions  &  15.1k &   4.7k &   1.3k \\
T2 & Somali news text       &  1.92M &  82.8k &  59.2k \\
T3 & Facebook posts         &  1.55M &  92.9k &  54.9k \\
T4 & Facebook comments      &   3.5M & 356.7k & 215.3k \\
T5 & LCC newspaper text     &  2.37M &   300k &   100k \\
T6 & LCC Wikipedia text     &   200k &  50.7k &    10k \\
T7 & LSTM generated text    & 11.29M &   4.7k & 775.3k \\ 
\bottomrule
\end{tabular*}
\vspace{-8pt}
\end{table}

\section{Semi-supervised training}
It has been shown that semi-supervised training can improve ASR performance in an under-resourced scenario \cite{yilmaz2018semi,nallasamy2012semi}. 
As we only have less than two hours of transcribed Somali acoustic data, increasing the pool of in-domain data by semi-supervised training was an attractive option.
To test this, we used a recently-acquired corpus comprising 17.55 hours of untranscribed Somali radio speech, as described in Section~\ref{sec:untranscribedacousticdata}. 
Since no speaker information was available, each utterance was considered to originate from a unique speaker. 

\begin{figure*} [h!]
	\centering
	\includegraphics[width=\textwidth ,height=2.8in]{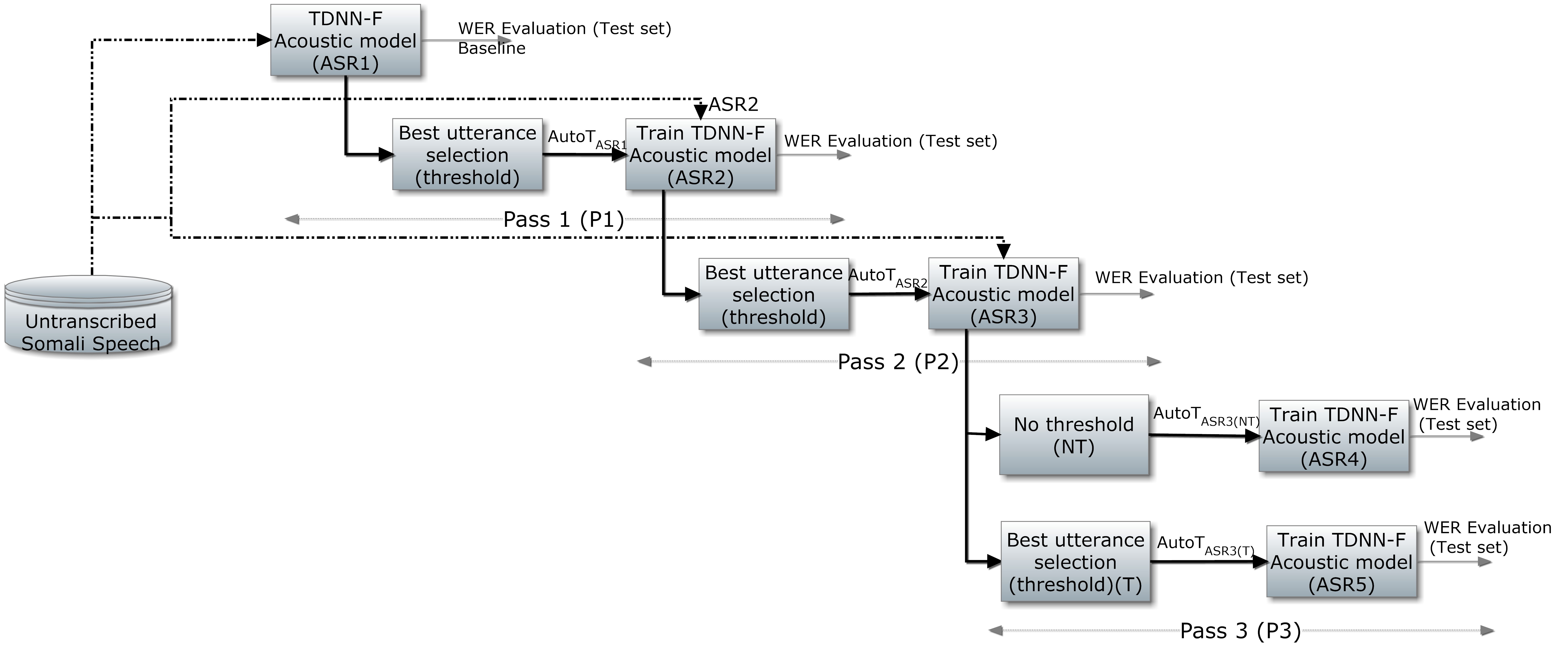}
	\caption{Semi-supervised training framework for Somali ASR. \protect\tikz[baseline]{\protect\draw[line width=0.3mm,dash dot, ->] (0,.8ex)--++(1,0)} represents untranscribed speech is being fed to transcriber}\label{mdnn}
		\label{fig_sst_scheme}
	\vspace{-12pt}
\end{figure*}

Our semi-supervised training strategy is shown in Figure~\ref{mdnn}.
The figure shows that we implemented three iterations of semi-supervised training, in each case re-transcribing the untranscribed data.
To start the process, we used our best previously-available Somali acoustic model~\cite{menon2018automatic} labelled ASR1 in Figure~\ref{mdnn}.

Because Somali is an agglutinative language, the number of unique word tokens is large, which poses challenges to ASR while decoding the utterances.
Thus, to improve the quality of the automatic transcriptions that are added to the training set, we applied a confidence threshold during each iteration.
The threshold was calculated as the average decoder confidence score across all automatically transcribed utterances.

By applying this procedure, 9.11 hours of the available 17.55 hours of untranscribed data was added to the training set in the first pass to develop ASR2.
In this second pass, 9.58 hours of automatically transcribed data was above the threshold and included in the training set to train ASR3.
 

Each training pass took approximately three days to complete on the limited available computational resources.
Hence, given this computational restriction and time constraints, we were not able to perform an exhaustive search for an optimum threshold in the earlier passes.
Nevertheless, in the final pass, we analysed the semi-supervised system performance for two different configurations:
\begin{itemize}
    \item Threshold = 0, meaning that the full 17.55 hours of semi-supervised data is considered (ASR4);
    \item Threshold = average decoder confidence, leading to 9.86 hours of semi-supervised data (ASR5).
\end{itemize}


Table~\ref{tab:trainingstrategies_and_results} describes the training strategies pass by pass for better insight.  



\section{Language modelling}
\label{sec:languagemodelling}
All language models were built using the SRILM toolkit \cite{stolcke2002srilm}.
The vocabulary of the ASR system was drawn from the pool of T1, T2 and T3 texts in Table~\ref{tab:textdata} by retaining all word types occurring at least four times.
The resulting vocabulary consisted of 41.7k word types.

The language model used in \cite{menon2018automatic} was used as the baseline \mbox{(LMbase).}
This model was trained on the Somali training set transcriptions (T1) and further interpolated with language models trained on the additional sources T2, T3, T4 and T7.
Sources T5 and T6 were not available at the time.
The interpolation weights were optimised on the Somali corpus test set (Table~\ref{tab:acousticdatasomali}).

Four additional language models were trained and evaluated.
The training and interpolation of these models was accomplished in much the same way as LMbase.
LM2 was trained on T1 and further interpolated with LMs trained on T2 to T7.
The interpolation weights were optimised on the test set.
LM3 was also trained on T1 and interpolated with LMs trained on T2 to T7, but interpolation weights were optimised on a held-out validation set extracted from the Somali acoustic training set transcriptions.
LM4 was trained on T1, interpolated with the LMs trained on T2 to T7 and further interpolated with the LM trained on the automatic transcriptions obtained at the output of ASR2 after the confidence threshold was applied.
The interpolation weights were optimised on the test set.
LM5 was also trained on T1, interpolated with the LMs trained on T2 to T7 and interpolated with the LM trained on the automatic transcriptions obtained at the output of ASR2 after the confidence threshold was applied.
However, in this case the interpolation weights were optimised on the held-out validation set.
The language model perplexities are shown in Table~\ref{tab:lms}.
The reason for using the AutoT transcriptions from the output of ASR2 is discussed later in Section~\ref{sec:results}.

\begin{table}[h]
\footnotesize
\centering
\renewcommand*{\arraystretch}{0.8}
\caption{Perplexities of the evaluated language models. (AutoT: Includes automatic transcriptions; PPval: Perplexity evaluated on the held-out validation set; PPtst: Perplexity evaluated on the test set; AutoT$_{ASR2}$: Automatic transcriptions from ASR2 after thresholding.)}
\vspace{-8pt}
\label{tab:lms}
\begin{tabular*}{0.47\textwidth}{@{\extracolsep{\fill}}llrr @{}}
\toprule
Language model & AutoT & PPval   &  PPtst \\
\midrule
LMbase         &     No &  --    &  269.80 \\
LM2            &     No &  --    &  253.60 \\
LM3            &     No & 576.98 &  321.31 \\
LM4            &    AutoT$_{ASR2}$ &   --   &  260.94 \\
LM5            &    AutoT$_{ASR2}$ & 500.49 &  300.25 \\
\bottomrule
\end{tabular*}
\vspace{-12pt}
\end{table}


\begin{table*}[t]
\footnotesize
\centering
\caption{
The various training configurations used for experimentation. ASR1 is the baseline system. (AM: Acoustic model; ManT: Hours of manually transcribed speech; AutoT: Hours of automatically transcribed speech; AutoT\textsubscript{ASRX}: Automatically transcribed speech obtained from ASR system `X'; LM: Language model; Superv.: Supervised; Semi: Semi-supervised; WER: Word error rate.)
}
\vspace{-6pt}
\label{tab:trainingstrategies_and_results}
\renewcommand{\arraystretch}{0.9}
\begin{tabular*}{0.85\textwidth}{@{\extracolsep{\fill}}lllrl@{}rlr@{}}
\toprule
\multirow{2}{*}{System} & \multirow{2}{*}{\parbox{5ex}{Training\newline strategy}}    & \multirow{2}{*}{AM} & \multicolumn{3}{c}{AM training data size (h)} & \multirow{2}{*}{LM} & \multirow{2}{*}{WER} \\
\cmidrule(lr){4-6}
                        &  &    &    ManT              &         \multicolumn{2}{c}{AutoT}             &                     \\
\midrule
Previous\cite{menon2018automatic} & Superv. & CNN-TDNN-BLSTM & 46.37 &  & 0.00 & LMbase & 53.75 \\
ASR1     & Superv. & TDNN-F         & 46.37 & & 0.00 & LMbase & 53.68 \\
ASR2     & Semi    & TDNN-F         & 46.37 & AutoT\textsubscript{ASR1}: &  9.11 & LMbase & 51.91 \\
ASR3     & Semi    & TDNN-F         & 46.37 & AutoT\textsubscript{ASR2}: &  9.58 & LMbase & \textbf{50.95} \\
ASR4     & Semi    & TDNN-F         & 46.37 & AutoT\textsubscript{ASR3}: & 17.55 & LMbase & 51.71 \\
ASR5     & Semi    & TDNN-F         & 46.37 & AutoT\textsubscript{ASR3}: &  9.86 & LMbase & 51.09 \\
\bottomrule
\end{tabular*}
\vspace{-8pt}
\end{table*}

\section{Acoustic modelling}
The Kaldi speech recognition toolkit was used for all ASR experiments~\cite{povey2011kaldi}.
All experiments were performed using a PC with an 8-core Intel i7 CPU, 32GB of RAM and a 12GB NVIDIA Tesla GPU.
In our previous work, we found multilingual training to improve ASR performance substantially~\cite{menon2018automatic}.
For multilingual training, the training sets of the four languages in Tables~\ref{tab:acousticdatasomali} and~\ref{tab:acousticdataother} were combined.
The lexica were concatenated and the phoneme labels were left unaltered, resulting in a combined, language-dependent phoneme set.
All ASR experiments used a closed vocabulary, i.e.\ no out-of-vocabulary words occur in the test data.
First, a  context-dependent GMM-HMM acoustic model with 25k Gaussians was trained using all the multilingual ManT data.
39-dimensional Mel-frequency cepstral coefficients (MFCC) with deltas and delta-deltas were used as features.
This GMM-HMM provided the alignments required for neural network training.
The same multilingual training data was used to compute acoustic features for neural network training.
However, in this case three-fold data augmentation was applied prior to feature extraction \cite{ko2015audio} and the acoustic features comprised 40-dimensional MFCCs (without derivatives), 3-dimensional pitch features and 100-dimensional i-vectors for speaker adaptation.

It has recently been shown that, when semi-orthogonal low-rank matrix factorisation is applied to the parameter matrices of TDNN layers, ASR performance can be improved in low-resource situations \cite{povey2018}.
Consequently, a TDNN-F acoustic model (10 time-delay layers followed by a rank reduction layer) was trained using the Librispeech recipe for Kaldi (version 5.2.164).
The factorisation decomposes the high-dimensional parameter matrix into two factor matrices, one of which is constrained to be semi-orthogonal.
This results in an intermediate bottleneck layer in the TDNN layer that has a lower dimension than the hidden layer.
This factorisation allows the TDNN-F model to use fewer parameters than hybrid architectures such as TDNN-LSTM and TDNN-BLSTM (bidirectional LSTM).
Hence, the TDNN-F models are faster to train.

Our TDNN-F was trained using the lattice-free maximum mutual information objective criterion~\cite{povey2016purely}.
No parameter tuning was performed during neural network training and the default recipe parameters were used.

\section{Results and discussion}
\label{sec:results}
The ASR performance is reported in Table~\ref{tab:trainingstrategies_and_results} in terms of the word error rate (WER) for the various training approaches.
In comparison with our previous ASR system \cite{menon2018automatic}, the improvement afforded by TDNN-F is clear (rows 1 and 2).
Even though TDNN-F uses only half the number of parameters as CNN-TDNN-BLSTM, it is able to offer better performance.

Next we consider the performance of semi-supervised training.
Comparing ASR1 and ASR2, we see that the first pass of decoding and retraining decreases the WER by 3.30\% relative to the baseline.
Due to the agglutination property, Somali has a large vocabulary and more semi-supervised training data helps to improve performance.
The neural network is able to learn from the larger training set afforded by the additional semi-supervised data.
In the second pass, ASR3 yields a further relative improvement of 1.85\% over ASR2.
The third pass (ASR4 and ASR5) did not show any further WER improvement.
ASR5, that was trained on the thresholded automatically transcribed data, was not able to outperform its counterpart from the previous pass (ASR3).
However, it was able to perform better than ASR4 which was trained using the full set of automatically transcribed data.
The results show that acoustic model training does not gain from the semi-supervised data after the second pass, and that the inclusion of poorly transcribed data leads to a deterioration in recognition performance.

\begin{figure} [b]
	\vspace{-10pt}
	\centering
	\includegraphics[width=\columnwidth]{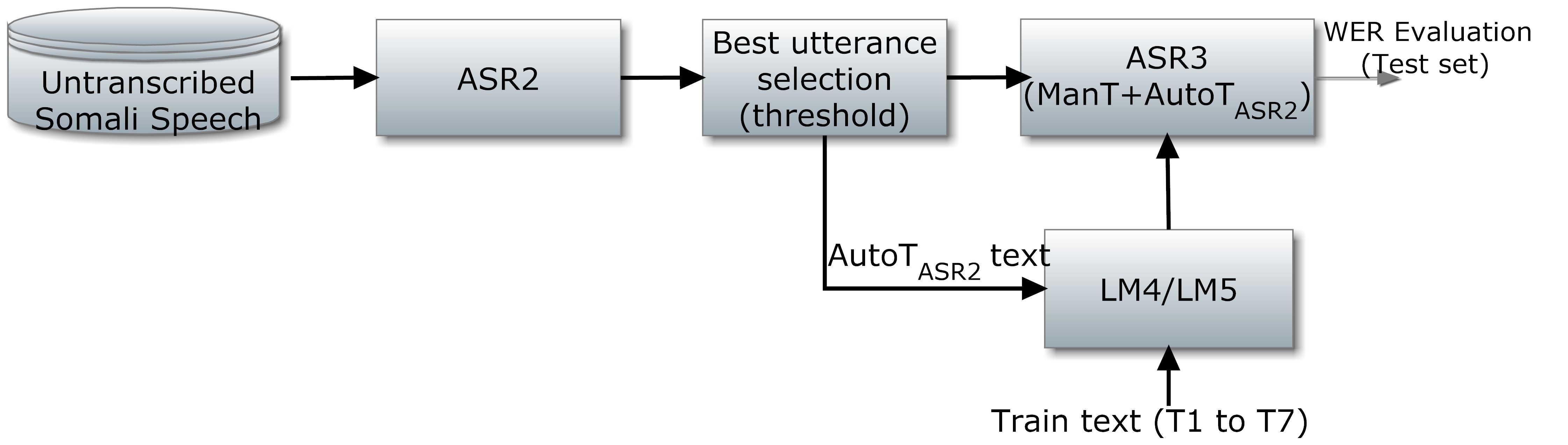}
	\caption{Semi-supervised acoustic and language modelling for Somali ASR3. }
		\label{fig_sst_lm}
\end{figure}

The best performing ASR system so far is ASR3, which was trained on the semi-supervised data produced by ASR2.
Next, we evaluate the language models described in Section~\ref{sec:languagemodelling} using the acoustic models of ASR3.
LM2 and LM3 were trained on the sources in Table~\ref{tab:textdata}, while LM4 and LM5 included the automatically transcribed text from ASR2 (AutoT\textsubscript{ASR2}) as additional training data.
Figure \ref{fig_sst_lm} gives a representation of the semi-supervised experimental framework for the language model evaluations.
LM2, which was optimised on the text set, did not show any significant improvement over the baseline.
However, LM3, which was optimised on the validation set, showed an improvement of 1.86\% relative to the baseline.
The results in Table \ref{tab:wer} show that the language models that included automatically transcribed text (LM4 and LM5) improve ASR performance.
The perplexity of LM5 was higher than the baseline, but it was optimised on the validation set and provided the better ASR performance, decreasing the WER by 3.32\%.
Overall, we achieved a 7.74\% relative improvement over our supervised baseline Somali ASR system, despite taking a simple approach to segmentation of the untranscribed speech.

\begin{table}[h]
\vspace{-2pt}
\centering
\caption{WER results of ASR3 with various language models.}
\label{tab:wer}
\resizebox{\columnwidth}{!}{%
\begin{tabular}{cccccccc}
\hline
Acoustic Model & LMbase & LM2 & LM3 & LM4 & LM5 \\ \hline
\begin{tabular}[c]{@{}c@{}}ManT Speech+ AutoT\textsubscript{ASR2} Speech\\ (TDNN-F)\end{tabular} & 50.95 & 50.89 & 50.00 & 50.20 & \textbf{49.59} \\ \hline
\end{tabular}%
}
\end{table}

\vspace{-10pt}
\section{Conclusion}
We have presented our initial efforts to increase the pool of Somali acoustic and language model data in a semi-supervised manner in an effort to improve automatic speech recognition for Somali.
A training corpus of only 1.57 hours of in-domain segmented and transcribed Somali radio broadcast speech data was available. 
A further 17.55 hours of unannotated Somali speech was segmented using a straightforward approach.
In addition, approximately 44.8 hours of annotated speech in three unrelated languages was available for multilingual modelling.
A baseline TDNN-F acoustic model was trained using this multilingual data and semi-supervised training was carried out in three passes.
In each pass, a threshold was applied to the confidence score of the decoded output, discarding utterances below the threshold.
We found that only the first two such passes of semi-supervised learning improved performance.
Although the segmentation approach for the untranscribed data was extremely simple, a 7.74\% relative improvement over the baseline system was achieved.
Ongoing work is exploring the effects of more sophisticated segmentation and diarisation approaches, as well as a more careful optimisation of the confidence threshold.

\section{Acknowledgements}
We thank the NVIDIA corporation for the donation of GPU equipment used for this research. We also gratefully acknowledge the support of Telkom South Africa.

\bibliographystyle{IEEEtran}
\bibliography{mybib}
\end{document}